\documentclass[letterpaper, 10 pt, conference]{ieeeconf}  % Comment this line out if you need a4paper

\IEEEoverridecommandlockouts                              % This command is only needed if 
\usepackage{listings}
\lstdefinestyle{FileFormat}{
  tabsize=1,
  showspaces=false,
  showstringspaces=false,
  xleftmargin=-\parindent,
  xrightmargin=.25in
}

\usepackage{epsfig}
\usepackage{amsmath} 
\usepackage{ amssymb }
\usepackage{url} 
\usepackage{xspace} 
\usepackage{graphicx} 
\usepackage{float} 
\usepackage{comment} 
\usepackage{tabu}
\usepackage[table]{xcolor}
\usepackage{color}
\definecolor{darkgreen}{rgb}{0,.75,0} 
\definecolor{tableShade}{gray}{0.95}

\title{SLAMBench2: Multi-Objective Head-to-Head Benchmarking\\ for Visual SLAM} 
\def\edinburgh{$^{\dagger}$}
\def\imperial{$^{\ddagger}$}
\def\manchester{$^{\star}$}
\def\stanford{$^{\bullet}$}
 \author{%
   Bruno Bodin\edinburgh{},
   Harry Wagstaff\edinburgh{},
   Sajad Saeedi\imperial{},
   Luigi Nardi\stanford{},
   Emanuele Vespa\imperial{},
   John Mawer\manchester{},
   Andy Nisbet\manchester{},\\
   Mikel Luj\'an\manchester{},
   Steve Furber\manchester{},
   Andrew J. Davison\imperial{},
   Paul H. J. Kelly\imperial{},
   and Michael F. P. O'Boyle\edinburgh{},  
   \vspace{-8 mm}
   \thanks{%
     \edinburgh{}~School of Informatics, University of Edinburgh, UK
   }
   \thanks{%
     \imperial{}~Department of Computing, Imperial College London, UK
   }
   \thanks{%
     \manchester{}~School of Computer Science, University of Manchester, UK
   }
  \thanks{%
    \stanford{}~Electrical Engineering - Computer Systems, Stanford University, USA
   }  
 }

\begin{document} 

\maketitle

\begin{abstract} 
SLAM is becoming a key component of robotics and augmented reality (AR) systems. While a large number of SLAM algorithms have been presented, there has been little effort to unify the interface of such algorithms, or to perform a holistic 
%large-scale 
comparison of their capabilities. This is a problem since different SLAM applications can have different functional and non-functional requirements. For example, a mobile phone-based AR application has a tight energy budget, while a UAV navigation system usually requires high accuracy.
SLAMBench2 is a benchmarking framework to evaluate existing and future SLAM systems, both open and close source, over an extensible list of datasets, while using a comparable and clearly specified list of performance metrics. A wide variety of existing SLAM algorithms and datasets is supported, e.g. ElasticFusion, InfiniTAM, ORB-SLAM2, OKVIS, and integrating new ones is straightforward and clearly specified by the framework. SLAMBench2 is a publicly-available  software  framework  which
represents  a  starting  point  for  quantitative,  comparable  and
validatable  experimental  research  to  investigate  trade-offs 
%in performance,  accuracy  and  energy  consumption  
across SLAM  systems.
%can thus be used as a tool to perform reproducible experiments for SLAM systems, and for the development and evaluation of new SLAM systems.  
\end{abstract}

\section{Introduction}

% The problem we want to solve: 
Over the last decade there has been sustained research interest in Simultaneous Localisation and Mapping (SLAM).
Every  year, new SLAM systems are released, each containing new ideas,  supporting new sensors, and tackling more challenging datasets.
As the area matures and the number of systems increases, the need to objectively evaluate them against prior work intensifies. 
However, robust quantitative, comparable and validatable evaluation of competitive approaches  
can take considerable effort, delaying innovation.

One of the main difficulties in robust evaluation is the numerous different
software platforms in which research is embedded. Each has to be installed and setup with the appropriate libraries and device drivers.
%More important is the fact that each system is typically aimed at a particular application context such as indoors, UAVs, or autonomous cars. This means that there will be particular sensors and datasets for which the system is designed. Applying the same system to different datasets may lead to failure. 
More important is the variations in input formats, hardware setups, and various parameters such as calibration and algorithmic parameters. These variations must be carefully managed to provide the desired results. For instance, applying similar parameters to different datasets may lead to failure. Trying to compare two different algorithms 
with the same sensors, datasets, and hardware platform may involve a significant amount of redundant software engineering effort.

%%It is not just the academic community which is concerned with robust evaluation.
%%Within industry a comparison of the state of the art is needed to select the best candidate in a particular environment, or to select the type of sensor required to reach the level of accuracy needed.

In this paper we present the SLAMBench2 benchmarking framework~\footnote{https://github.com/pamela-project/slambench2}, which aims to provide reproducibility of results for different datasets and SLAM algorithms.
It provides a single plug and play environment where users can try different algorithms, sensors, datasets, and evaluation metrics, allowing robust evaluation.

The need to systematise and reduce redundant evaluation effort  is not new and has already been recognised, e.g. the SLAM Evaluation 2012 from the KITTI Vision Benchmark Suite~\cite{Geiger2012CVPR}
%This work provided a meaningful snapshot of the state of the art, although it is 
which is limited to a specific dataset and environment.
More recently, SLAMBench~\cite{Nardi2015ICRA} provided multiple implementations (C++, OpenMP, OpenCL and CUDA) of a single algorithm, KinectFusion,
and dataset, ICL-NUIM~\cite{Handa2014ICRA}.
%with trajectory and scene ground-truth for reliable trajectory accuracy %comparison of those different implementations of the algorithm.

In this paper we introduce SLAMBench2, which is designed to enable the evaluation of %existing and future 
open source or proprietary
SLAM systems,  
over an extensible sets of datasets and clearly specified metrics. SLAMBench2 is inspired by the work in \cite{Nardi2015ICRA,zia2016comparative,nardi2017algorithmic}.
It currently supports eight different algorithms (dense, semi-dense, and sparse) and three datasets.
It is a tool that enables the reproducibility of results for existing SLAM systems and allows the integration and evaluation of new SLAM results.
Through a series of use-cases (Sec.~\ref{sec:experiments}), we demonstrate its simplicity of use as well as a variety of results that have already been collected.
We make the following contributions: 
\begin{itemize}
\item A publicly-available framework that enables quantitative reproducible comparison   of SLAM  systems.
\item A system that is {\em dataset-agnostic}, supports {\em plug and play} of new algorithms, and provides a 
{\em modular user interface}.
\item A quantitative comparison between eight state-of-the-art SLAM systems.
\end{itemize}

\section{Background and Related Work} 
%%Missing Citations
\subsection{Simultaneous Localisation And Mapping}
Simultaneous Localisation and Mapping is the process of localising a sensor in a previously unexplored environment while building a map of its environment incrementally.
The interdependence of localisation and mapping makes the problem more complex, and researchers have proposed various solutions such as joint optimisation using Kalman filters~\cite{Thrun:2005} or parallel tracking and mapping~\cite{klein2007parallel}.

While in theory one may claim that SLAM is a solved problem \cite{Frese:2010:Springer}, in practice it is still considered challenging  \cite{Cadena:IEEETRO:2016} when a specific level of performance is required (e.g. frame rate, accuracy).
For example, in visual SLAM, when a camera moves rapidly or the environment is highly dynamic, SLAM may simply fail to produce consistent maps.
Therefore, improving the performance and robustness of SLAM algorithms and their applications is still an important research topic~\cite{Cadena:IEEETRO:2016,SAEEDI:JFR:2016,vespa2018efficient}.

In visual SLAM, various discrete paradigms exist for the algorithms.
For instance, {\bf `direct'} algorithms use all pixels' intensity values directly~\cite{2011Newcombe}, while {\bf `indirect'} methods extract features and process those~\cite{davison2007monoslam}. 
Orthogonal to the direct/indirect paradigm is the reconstruction spectrum~\cite{DSO2018Engel}. On one end of the spectrum are {\bf `sparse'} methods, where only a subset of features are used for reconstruction~\cite{davison2007monoslam}. 
These methods, also called feature-based, tend to use less resources in terms of power and system memory. 
Sparse methods limit the amount of information extracted from images. However, they may be useful in situations where a detailed map is not required. 
On the other end are {\bf `dense'} methods  where all pixels are used for reconstruction~\cite{Whelan2015RSS}. Additionally, in sparse method points are treated independently, while in dense methods, geometric dependencies are used. Semi-dense algorithm~\cite{engel2014lsd} belong to the middle of this spectrum.
Within each category, various algorithms exist, depending on the details of the design.
For example, in sparse visual SLAM, ORB-SLAM~\cite{Artal:IEEETRO:2015} uses ORB features~\cite{Rublee:ICCV:2011} and OKVIS~\cite{leutenegger2015keyframe} relies on BRISK features~\cite{Leutenegger:ICCV:2011}.
This diversity makes evaluation and comparison difficult.

In SLAMBench2, we support several different algorithms. 
Amongst sparse algorithms, MonoSLAM~\cite{davison2007monoslam}, PTAM~\cite{klein2007parallel}, OKVIS~\cite{leutenegger2015keyframe}, and ORB-SLAM2~\cite{Artal:IEEETRO:2015} are supported. The first two algorithms support only monocular systems, 
OKVIS supports only stereo and RGB-D, and ORBSLAM2 support all. These algorithms are indirect algorithms. 
Amongst dense methods, three algorithms are supported. The previous release of SLAMBench integrated an implementation of KinectFusion~\cite{2011Newcombe}, a dense algorithm only using depth frames to produce its map.  
In SLAMBench2 in addition to KinectFusion, we also integrated ElasticFusion~\cite{Whelan2015RSS} and InfiniTAM~\cite{InfiniTAM_ISMAR_2015}, two recent dense systems with fundamentally different data structures and operators than KinectFusion.
Finally LSD-SLAM, a semi-dense SLAM system using only one monocular camera, is also supported in SLAMBench2. These algorithms are summarised in Table~\ref{fig:algorithms}. 

\subsection{Computer Vision and Robotics Middleware}
With new algorithms being proposed, it is clear that SLAM designers need to compare different algorithms efficiently and systematically. Additionally, with this large variety of algorithmic choices, it is a key concern for SLAM users how to compare different algorithms/parameters, and select the most appropriate. SLAMBench2 presents a platform that researchers in industry and academia can harness to efficiently and fairly compare different algorithms.
Within the robotics and computer vision community, the closest related works to SLAMBench2 are various types of middleware such as Robot Operating System (ROS)~\cite{Quigley2009ICRA}, Gazebo~\cite{Hsu2012ROSCON}, Orca~\cite{Orca}, OpenRDK~\cite{Calisi2008ICIRS}, YARP~\cite{Metta2006IJARS}, MOOS~\cite{Newman2008moos}, and many more. These systems have been designed to facilitate design, simulation, and integration of sensors, algorithms, and actuators. SLAMBench2 is not a robotic middleware, but rather it is a benchmarking framework with generic APIs concerned about SLAM and computer vision algorithms. SLAMBench2 aims to help designers and developers to design and compare their algorithms against the state-of-the-art with minimum overhead. In addition, this enables fair comparison between different hardware to be performed on many state-of-the-art SLAM algorithms.

\subsection{SLAM Benchmarks}
Our work is closely related to KITTI Benchmark Suite~\cite{Geiger2012CVPR} and TUM RGB-D benchmarking~\cite{Sturm2012IROS}.
However we differentiate from these systems by expanding in two directions: algorithms and datasets.
SLAMBench2 allows users to evaluate  a new trajectory, real-world or synthetic, against existing algorithms available in the platform.
Additionally, the diversity of the datasets allows researchers to evaluate new performance metrics such as reconstruction error and reconstruction completeness.

\section{SLAMBench2}

\begin{figure}
	\centering
	\includegraphics[width=\linewidth]{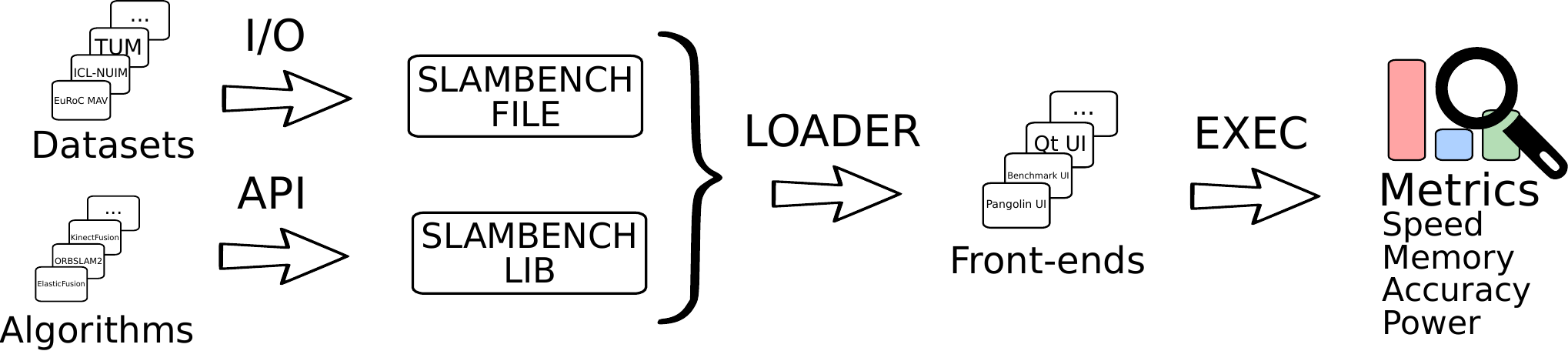}
	\caption{Overview of the SLAMBench2 structure. This picture shows the four main components of the framework (I/O, API, Loader, and User Interface) as well as its three main characteristics (Dataset-agnostic, Plug and Play Algorithms, and Modular User Interfaces).}
	\label{fig:slambench_overview}
\end{figure}

\subsection{Overview}

SLAMBench2 goes beyond previous frameworks~\cite{Nardi2015ICRA} in that it is
{\em dataset-agnostic}, supports {\em plug and play algorithms}, and provides
{\em modular user interfaces}.
See Fig.~\ref{fig:slambench_overview} for a summary.

\noindent \textbf{Dataset-agnostic}:
It includes interfaces to use ICL-NUIM, TUM RGB-D, and EuRoC MAV (described in Sec.~\ref{sec:datasets}), and can be extended to any dataset using  compatible sensors.
Datasets consisting of new sensors can also be supported by extending the SLAMBench2 library.
\\

\noindent \textbf{Plug and Play Algorithms}:
It provides a generic API to interface with SLAM algorithms and  integrate new SLAM algorithms.
There are currently eight different SLAM systems available (described in Sec.~\ref{sec:slam_algorithms}).
SLAMBench2 also works with  closed source SLAM systems which can  be distributed to the community without sharing source code.
This feature is expected to raise the level of collaboration with industry in SLAM research.
\\

\noindent \textbf{Modular User Interface}:
SLAMBench2 also makes it easy to select between different user interface systems.
Evaluation metrics can be changed and customised, as can the Graphical User Interface (GUI), while maintaining independence of the datasets and SLAM algorithms.
For instance, the native visualiser, based on the Pangolin library~\cite{Pangolin}, can easily be replaced with ROS visualiser~\cite{RViz}.
Although this modularity is useful for advanced evaluations, we provide and recommend default user interfaces to maximise the reproducibility of experimental results.

\subsection{Framework Components}

The framework has four main components:  \textbf{I/O System} (Sec.~\ref{sec:io}),  \textbf{Integration API} (Sec.~\ref{sec:api}), \textbf{Loader} (Sec.~\ref{sec:loader}) and \textbf{User Interfaces} (Sec.~\ref{sec:loader}).
\begin{enumerate}
\item The I/O System translates existing datasets into SLAMBench2 datafiles which contain a description of the dataset sensors, frames, and ground-truth data when available (trajectory and 3D point cloud). For calibration, currently pinhole camera model with radial distortion is supported. The model includes two parameters for focal length and two parameters for optical center along $x$ and $y$ axes, and five parameters for lens distortion following the OpenCV format.
\item The API is used to interface to any SLAM algorithm implementation, and includes features to:
\begin{itemize}
\item describe algorithmic parameters,
\item stream frames into the algorithm,
\item collect results (camera pose and reconstruction).
\end{itemize}
\item The Loader is then used to load the datafile and 
SLAM algorithm using the API. The loader executes these SLAM algorithms using the specified dataset.
\item Finally User Interfaces are used to dynamically analyse the algorithm; these can be command-line tools for accurate experiments, or GUI for demonstrations and debugging.
\end{enumerate}
 
\subsection{Supported Artefacts}

\subsubsection{Datasets}
\label{sec:datasets}

SLAMBench2 supports a number of existing datasets incorporating a range of different sensors and sensor configurations (listed in Table. \ref{fig:datasets}).
Due to the presence or absence of certain types of sensor in each dataset, most of the datasets are compatible only with a subset of the supported algorithms.

\begin{table}
  \small
  \begin{center}
    \renewcommand{\arraystretch}{0.96}
    \begin{tabular}{|l|l|l|}
      
				\hline				
                                \multicolumn{3}{|c|}{\textbf{ICL-NUIM Dataset}~\cite{Handa2014ICRA}}  \\
                                \hline
                                \multicolumn{3}{|l|}{\small~Sensors: RGB-D}\\
                                \multicolumn{3}{|l|}{\small~Ground-truth: Trajectory, 3D Point Cloud}\\
				\hline
		    		Name & Duration & datafile size \\
				\hline
				Living Room 0 &   60.4s     & 2.8GB   \\
				Living Room 1 &   38.6s     & 1.8GB \\
				Living Room 2 &   35.2s     & 1.7GB \\
				Living Room 3 &   49.6s     & 2.3GB  \\
				\hline
				\multicolumn{3}{|c|}{\textbf{TUM-RGDB Dataset}~\cite{Sturm2012IROS}}  \\
                                \hline
                                \multicolumn{3}{|l|}{\small~Sensors: RGB-D, IMU}\\
                                \multicolumn{3}{|l|}{\small~Ground-truth: Trajectory}\\
				\hline
				Name & Duration & datafile size \\
				\hline
				Freiburg1/xyz    & 30.1s  & 1.4GB \\
				Freiburg1/rpy    & 27.7s  & 1.3GB \\
				Freiburg1/360    & 28.7s  & 1.3GB \\
				Freiburg1/floor  & 49.9s  & 2.2GB \\
				Freiburg1/desk   & 23.4s  & 1.0GB \\
				Freiburg1/desk2  & 24.9s  & 1.1GB \\
				Freiburg1/room   & 48.9s  & 2.4GB \\
				Freiburg2/xyz    & 122.7s & 6.6GB \\
				Freiburg2/rpy    & 110s   & 5.9GB \\
				Freiburg2/360\_hemisphere    & 91.8s       & 4.9GB\\
				Freiburg2/360\_kidnap        & 48s       & 2.5GB\\
				Freiburg2/desk               & 99.4s       & 5.3GB \\
				Freiburg2/desk\_with\_person & 142.1s       & 7.3GB \\
				Freiburg2/large\_no\_loop    & 112.4s       & 6.0GB \\
				Freiburg2/ large\_with\_loop & 173.2s       & 9.3GB \\
				Freiburg2/pioneer\_360       & 72.8s       & 2.1GB \\
				Freiburg2/pioneer\_slam      & 155.7s       & 5.2GB \\
				Freiburg2/pioneer\_slam2     & 115.6s       & 3.7GB \\
				Freiburg2/pioneer\_slam3     & 111.9s       & 4.5GB \\
				\hline
				\multicolumn{3}{|c|}{\textbf{EuRoC MAV Dataset}~\cite{Burri2016IJRR}}  \\
                                \hline
                                \multicolumn{3}{|l|}{\small~Sensors: Greyscale Stereo, IMU}\\
                                \multicolumn{3}{|l|}{\small~Ground-truth: Trajectory, 3D Point Cloud}\\
				\hline
				Name & Duration & datafile size \\
				\hline
                                Machine Hall 1 & 184.1s  & 10.3GB \\
                                Machine Hall 2 & 152s    &  8.5GB \\
                                Machine Hall 3 & 135s    &  7.6GB \\
                                Machine Hall 4 & 101.6s  &  5.7GB \\
                                Machine Hall 5 & 113.6s  &  6.4GB \\
                                Vicon Room 1/1 & 145.6s  &  8.2GB \\
                                Vicon Room 1/2 & 85.5s   &  4.8GB \\
                                Vicon Room 1/3 & 107.5s  &  6.0GB \\
                                Vicon Room 2/1 & 114s    &  6.4GB \\
                                Vicon Room 2/2 & 117.4   &  6.6GB \\
                                Vicon Room 2/3 & 116.9s  &  6.0GB \\
				\hline
			\end{tabular}
    \caption{
                          Compatible datasets with the SLAMBench2 framework.
                          The datafile size is the uncompressed SLAMBench datafile.}                        
		\label{fig:datasets}
  \end{center}
  \vspace{-2em}
	\end{table}

\subsubsection{SLAM Algorithms} 
\label{sec:slam_algorithms}

There are many existing SLAM systems and it is likely that  future systems will considerably improve on the current state-of-the-art.
Therefore an important feature is the easy integration of new SLAM systems into the framework.
SLAMBench2 already has eight different algorithms fully integrated, which can be used as examples for the integration of new algorithms (See Table ~\ref{fig:algorithms}).

\begin{table*}
	\centering
	\begin{tabular}{l l l l c c l}
		\textbf{Algorithm} & \textbf{Type} & \textbf{Sensors} & \textbf{Implementations} & \textbf{Large-scale} & \textbf{Loop-closure} & \textbf{Year} \\ 
\hline
		ORB-SLAM2~\cite{murORB2} & Sparse &  RGB-D, Monocular, Stereo& C++ & \checkmark & \checkmark & 2016\\
		MonoSLAM~\cite{davison2007monoslam} & Sparse & Monocular & C++, OpenCL & \checkmark &  & 2003\\
		OKVIS~\cite{leutenegger2015keyframe} & Sparse & Stereo, IMU & C++ & \checkmark &  & 2015\\
		PTAM~\cite{klein2007parallel} & Sparse & Monocular & C++ & \checkmark &  & 2007\\        
		\hline
		ElasticFusion~\cite{Whelan2015RSS} & Dense & RGB-D & CUDA & \checkmark & \checkmark & 2015\\
		InfiniTAM~\cite{InfiniTAM_ISMAR_2015} & Dense &  RGB-D & C++, OpenMP, CUDA & \checkmark &  & 2015\\
		KinectFusion~\cite{2011Newcombe} & Dense & RGB-D & C++, OpenMP, OpenCL, CUDA & &   & 2011\\
		\hline
		LSD-SLAM~\cite{engel2014lsd} & Semi-Dense & Monocular & C++, PThread & \checkmark & \checkmark & 2014\\
        \hline
	\end{tabular}
	\caption{
          Table summarising the algorithms currently adapted for use in SLAMBench2. We mark with a \checkmark when the functionality is included. The year column shows when the algorithm appeared for the first time.
        }
        \label{fig:algorithms}
        \vspace{-2em}
\end{table*}

\subsection{Performance Metrics}

The original release of SLAMBench had a number of features designed around providing various metrics: frame rate (measured in frames per second), absolute trajectory error (ATE, measured in centimetres, explained in Section~\ref{sssec:accuracy}), and power consumption (measured in Watts).
SLAMBench2 provides these alongside several new metrics.
We also extend some of the existing metrics to provide more information, and to be compatible with more experimental configurations (e.g. devices).

\subsubsection{Computation Speed}

Firstly, the computation speed metric provided in the original SLAMBench is preserved. For each algorithm, the total computation time per frame is available. It is also possible for an integrated implementation to provide more fine-grained timing information. For example, times for the front-end image processing, tracking, and integration are all available from KinectFusion. Many of the currently integrated algorithms provide these fine-grained metrics, which can be used to tune parameters or implementations in order to improve performance.

\subsubsection{Algorithmic Accuracy} \label{sssec:accuracy}

The accuracy metrics are determined by comparing estimates with ground-truth data. Absolute Trajectory Error (ATE) and Relative Pose Error (RPE) are used to measure trajectory accuracy. RPE is a measure for the drift of the trajectory over a fixed time interval. While RPE measure the local accuracy of the estimate trajectory, ATE measures the global consistency of the trajectory. For more information about ATE and RPE, refer to~\cite{Sturm2012IROS}. These trajectory metrics alongside with a mapping metric, Reconstruction Error (RER)~\cite{Whelan2015RSS}, provide quantitative comparisons for various algorithms.

ATE and RPE are computed at runtime, with a minimal alignment between the first closest ground-truth pose and estimated pose (in terms of timestamp).
Off-line, more complex alignment techniques can be used in order to compare dense and semi-dense techniques when the map scales do not match. 
RER is determined by running iterative closest point (ICP) algorithm on the point cloud models of the reconstruction and ground truth.
As this is computationally intense, this evaluation is also performed off-line.

\subsubsection{Power Consumption}

In the original SLAMBench \cite{Nardi2015ICRA}, the power consumption metric was relatively ad-hoc, and only supported a specific embedded system (ODROID-XU3).
We extended the metric  to support PAPI~\cite{Mucci1999HPCMP} (a generic power sensor API).
In the future we hope to provide more extensive power consumption metrics.
However, fine-grained power sensors are still uncommon in embedded development boards, and tend to have an ad-hoc or vendor-specific interface, which makes integration  challenging.

\subsubsection{Memory Usage}

Finally we include memory consumption as a new metric. The dynamic memory consumption of an algorithm can be accurately measured at a per-frame level, and overall memory consumption can also be reported. We consider this to be an important contribution, as SLAM algorithms tend to consume a large amount of memory and memory consumption is often not considered as a goal or evaluated. For implementations that use GPU acceleration, we also provide methods to measure how much on-GPU memory is consumed. Due to the multiple GPU compute APIs and GPU vendors, these techniques are ad-hoc, but can be used to provide information on the relative memory use of each algorithm and dataset.

The next sections will detail the different components of SLAMBench2, these are the I/O system, the API, the loader and user interfaces. 

\section{SLAMBench2 Core Components}
\subsection{I/O System}
\label{sec:io}

The I/O system is used to prepare datasets to be used as inputs to the SLAMBench2 framework.
As mentioned above, existing implementations of SLAM algorithms tend to each use their own dataset format.
This makes it very difficult to compare the performance of two algorithms on the same dataset, since the dataset would need to be made available in multiple different formats. 

SLAMBench2 attempts to solve this problem by defining a simple unified format, and then providing tools to convert from the dataset formats of popular datasets into this format.
Creating such conversion tools, to support new dataset formats, is straightforward.

The SLAMBench2 datafile is defined as follows:

\lstset{basicstyle=\footnotesize\ttfamily,style=FileFormat}
\begin{lstlisting}
  DATAFILE = <HEADER><SENSORS><GT_FRAMES><IN_FRAMES>
  HEADER   = <VERSION:4B><SENSOR_COUNT:4B>
  SENSORS  = <SENSOR 1>...<SENSOR N>
  SENSOR   = <TYPE:4B><PARAMETERS>
  GT_FRAMES= <EMPTY>|<GT_FRAME 1>...<GT_FRAME N>
  IN_FRAMES= <IN_FRAME 1>...<IN_FRAME N>
  GT_FRAME = <TIMESTAMP:8B><GT_TYPE:4B><DATA>
  IN_FRAME = <TIMESTAMP:8B><IN_TYPE:4B><DATA>
  GT_TYPE  = POSE|POINT_CLOUD
  IN_TYPE  = RGB_FRAME|DEPTH_FRAME|IMU_FRAME|...
\end{lstlisting}

The datafile contains a description of the sensors used within the dataset (\texttt{SENSORS}), followed by a (possibly empty) collection of `ground-truth' frames ordered by timestamp (\texttt{GT\_FRAMES}), followed by a collection of input frames ordered by timestamp (\texttt{IN\_FRAMES}).
The data files can contain multiple sensors of the same type.
For example, multiple RGB camera sensors can be used to make up a stereoscopic sensor.

A large variety of input frame types are supported, including: RGB, Greyscale Images, Depth Images, IMU Data, and Pixel Events.

In order to allow for measurements of trajectory and reconstruction error, the SLAMBench2 data format supports several `ground-truth' frame types. This allows a data file to contain both the input and expected output. Two such frame types are currently supported: Ground-Truth Trajectory data and Reconstruction Point Cloud data.

Image sensors present several problems due to differing image formats, pixel formats, etc. We choose to address this by using a single uncompressed raster image format with a configurable pixel format.
Although using an uncompressed format means that the data size is large, we do this to avoid the runtime costs involved in uncompressing the image data.
Calibration settings of cameras are also integrated in the datafile, though those parameters can be changed at runtime if needed.

Reading and writing dataset files is performed via a library provided with SLAMBench2. Several programs which can be used to convert between ICL-NUIM, TUM RGB-D, and EuRoC MAV formats are provided as examples of how to produce these files. Data is then read out of the file one frame at a time, with a variety of helper classes provided.

\subsection{API}
\label{sec:api}

One of the most critical functions of the SLAMBench2 infrastructure is to actually interface with implementations of SLAM algorithms.
Implementations of SLAM algorithms are often released as standalone executables (rather than as libraries), and often do not conform to any standard API or structure.
A possible approach would be to use a middleware such as ROS~\cite{Quigley2009ICRA}, but there is no clear standard defined to declare SLAM systems in this context and using such an infrastructure may add more constraints on the system to be used.
In contrast, SLAMBench2 provides a standardised API which is used as an interface between the SLAM algorithm implementation and other components of the SLAMBench2 infrastructure.
This API is used to initialise the data structures used by the implementation, send input data frames into the implementation, actually execute the algorithm, and to extract the outputs, i.e. trajectory and reconstruction.

SLAM algorithms are typically highly configurable, so that they can be adapted to different operating conditions and to different host systems, e.g. to gracefully degrade accuracy but improve speed in order to run on a less powerful device, or to gracefully degrade speed but improve power consumption for a power-constrained device. These parameters can be exposed through the SLAMBench2 API in order to allow user interfaces to manipulate these parameters. This also means that SLAMBench2 can be used in combination with auto-tuning techniques, this point will be the subject of one use-case in Sec.~\ref{sec:autotuning}.

\begin{figure}
	\centering
	\includegraphics[width=\linewidth]{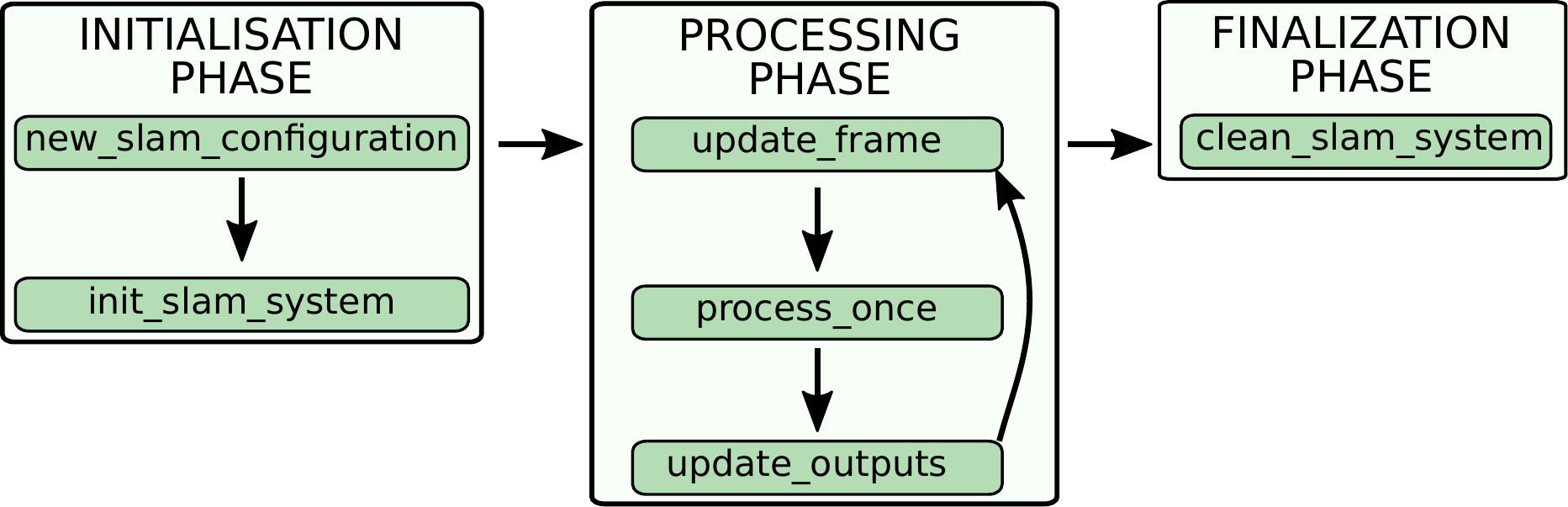}
	\caption{Workflow of the SLAMBench API.}
	\label{fig:slambench_api_workflow}
\end{figure}

The SLAMBench2 API is made up of the following functions:
  \begin{lstlisting}
    bool sb_new_slam_configuration(SBConfig*);
    bool sb_init_slam_system(SBConfig*);
    bool sb_update_frame(SBConfig*, SBFrame*);
    bool sb_process_once(SBConfig*);
    bool sb_update_outputs(SBConfig*);
    bool sb_clean_slam_system();
  \end{lstlisting}
This API is used by the framework through a 3-phase workflow as shown in Fig.~\ref{fig:slambench_api_workflow}.
In the rest of this section we will describe this workflow.

\subsubsection{Initialisation Phase}

The first functions of this API are used to initialize the SLAM system.

\begin{itemize}
\item \texttt{sb\_new\_slam\_configuration}:
  The \texttt{SBConfig} object is used to share environment information between the framework and the SLAM algorithm. The algorithm is also able to use this object to declare parameters.
Later, this information enables the GUI to show these parameters, to change them on-line, or to perform auto-tuning, which is optimising parameters value using an automatic mechanism for accuracy/performance trade-offs (e.g. \cite{Bodin2016PACT,nardi2017algorithmic}), or for active SLAM (e.g. \cite{saeedi2017application}).
\item \texttt{sb\_init\_slam\_system}:
  This function controls the start of memory allocation and SLAM system initialisation.
  This time the \texttt{SBConfig} object contains user-defined values for the algorithm parameters as well as a description of the current available sensors.
\end{itemize}

\subsubsection{Processing Phase}

For every frame, the \texttt{sb\_update\_frame} function is called to share frame data with the SLAM algorithm.
When the algorithm is capable of doing meaningful computation (where the data received are sufficient), then SLAMBench2 will call the \texttt{sb\_process\_once} function.\\
After calling \texttt{sb\_process\_once}, SLAMBench2 calls \texttt{sb\_update\_outputs} to get the current pose estimation, map estimation and tracking status from the SLAM system.
Extraction of the map can be complex given the wide range of possible formats, this is discussed later in Sec.~\ref{sec:map_extraction}.

\subsubsection{Finalisation Phase}

Eventually when the user interface is closed or when the end of data is reached, SLAMBench2 will call the \texttt{sb\_clean\_slam\_system} function.
In this function the SLAM system is expected to release any memory buffers previously allocated.

\subsubsection{Graphical Rendering and Map Extraction}
\label{sec:map_extraction}
As there are a large number of different data structures used to internally represent a map, we did not define a specific format to extract the map. Instead we defined an \texttt{OutputManager}, used to store any information from the SLAM system which could potentially be processed or visualised. This includes trajectories and maps as much as visual information for debugging purposes.
When SLAMBench2 calls \texttt{sb\_init\_slam\_system}, the SLAM system is made aware that the UI system is being used, and so memory allocation or other initialisation specific to the UI should be performed.
Then, after every call of \texttt{sb\_process\_once}, SLAMBench2 also calls the \texttt{sb\_update\_outputs} function which allows the SLAM system to update the information displayed on the user interface.
If the data structure used by the SLAM system is new, there is a way to include this new format.
For now, SLAMBench2 already defines several types such as a 3D point cloud, a list of visible features, and RGB frames.

\subsection{Loader and User Interfaces}
\label{sec:loader}
\label{sec:ui}
The Loader connects the user interface with algorithms. The Loader is a critical part of the infrastructure in the context of collaboration with industry, since it allows SLAM libraries to be dynamically loaded without source code being made available. This means that commercially-developed SLAM algorithms can be integrated into SLAMBench2 and compared against other commercial or open source algorithm implementations in a consistent and reproducible manner.

\begin{figure*}
  \centering
	\includegraphics[height=14em]{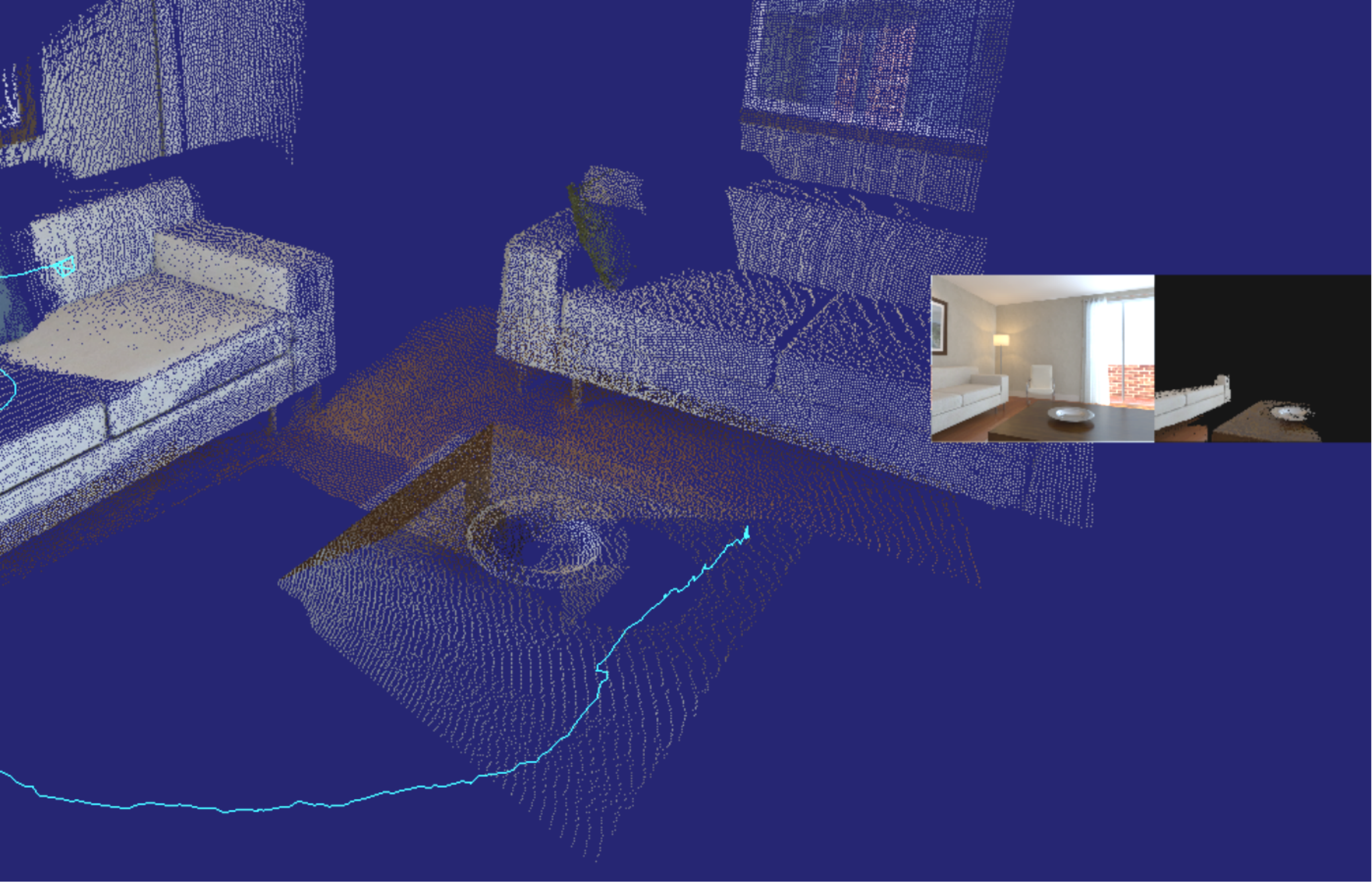}
	\includegraphics[height=14em]{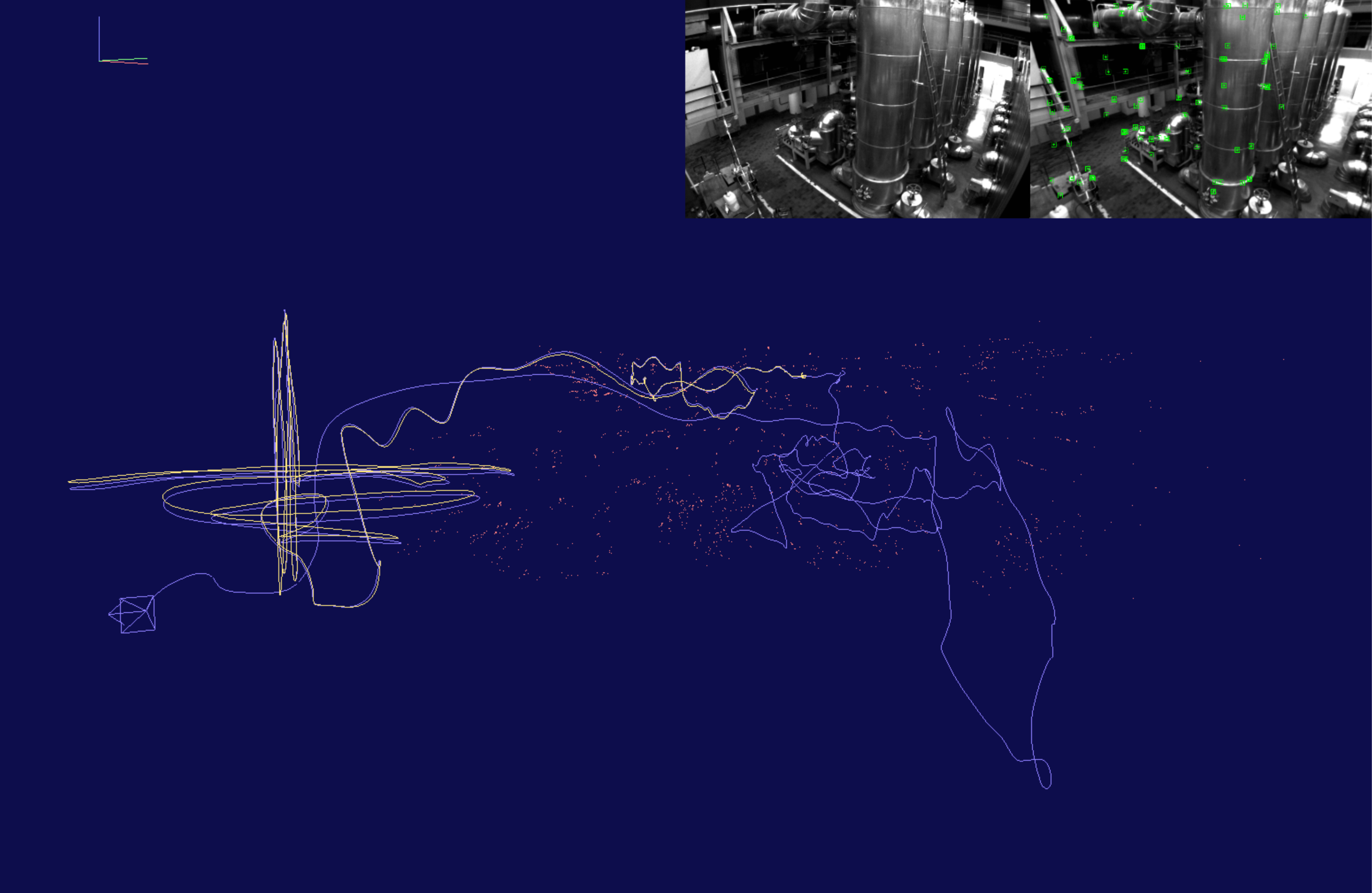}
	\caption{
          Cropped screenshots of the GUI of SLAMBench2 that uses the Pangolin library~\cite{Pangolin}.
          On the left a point cloud extracted from ElasticFusion is shown, alongside the trajectory.
          The input frame and internal 3D rendering of ElasticFusion are also shown, inlaid on the right of the image.
          On the right a similar interface used with ORBSLAM2 and the EuRoC MAV dataset. This time the map is sparse, we can see the trajectory aligned with the ground-truth.
        }
        \label{fig:screenshot}
\end{figure*}

The user interfaces display the outputs and metrics of the algorithms, such as output trajectories, accuracy, reconstruction, etc.
Multiple interfaces are available, with two of the main interfaces being a GUI which can be used to view and compare trajectories and reconstructions between different algorithms and ground-truths, and a `benchmark' interface, which has a text-only interface and outputs the desired metrics (such as timing or trajectory information).
Figure~\ref{fig:screenshot} shows the existing GUI.

\section{Experiments}

This section demonstrates how SLAMBench2 can be used for a range of 
use cases. Here. we evaluate  memory size, accuracy, speed and power consumption for different algorithms, datasets, and devices. Each of these results can be 
reproduced with a few command lines.

\subsection{Experimental Setup}
For our experiments we used three platforms, we will refer to these machines as ODROID, TK1, and GTX1080.

\textbf{ODROID}: A Hardkernel ODROID-XU3 based on the Samsung Exynos 5422 SoC, featuring an ARM big.LITTLE heterogeneous multiprocessing (HMP) unit with four Cortex-A15 ``big'' out-of-order cores, four Cortex-A7 ``LITTLE'' in-order cores, and a 6-core Mali-T628-MP6 GPU.  

\textbf{TK1}: A Jetson TK1, NVIDIA embedded Linux development platform featuring a Tegra K1 SoC. 
It is provided with a quad-core ARM Cortex-A15 (NVIDIA 4-Plus-1), with an on-die GPU composed of 192 CUDA cores.

\textbf{GTX1080}: At the desktop end of the spectrum we selected a machine featuring an Intel i7-6700K CPU and an NVIDIA GTX 1080 GPU containing 2560 CUDA cores.

\subsection{Experimental Evaluation} 
\label{sec:experiments}

\subsubsection{Memory-Constrained Device}

While memory consumption of SLAM systems is not typically considered, 
there are areas  where available  memory is highly constrained.
For example, Augmented Reality applications that run on mobile phones often have only 4GB of RAM available, which must be shared with the operating system and other apps. We therefore evaluated a variety of SLAM algorithms and measured how  memory usage  varies throughout the dataset. For edge computing platforms the memory constraint is even more important allowing only few hundred of Megabytes of RAM. By introducing this performance metric in the design space SLAMBench2 pushes the envelope of deployability benchmarking of SLAM systems. 

\begin{figure*}
  \includegraphics[width=\linewidth]{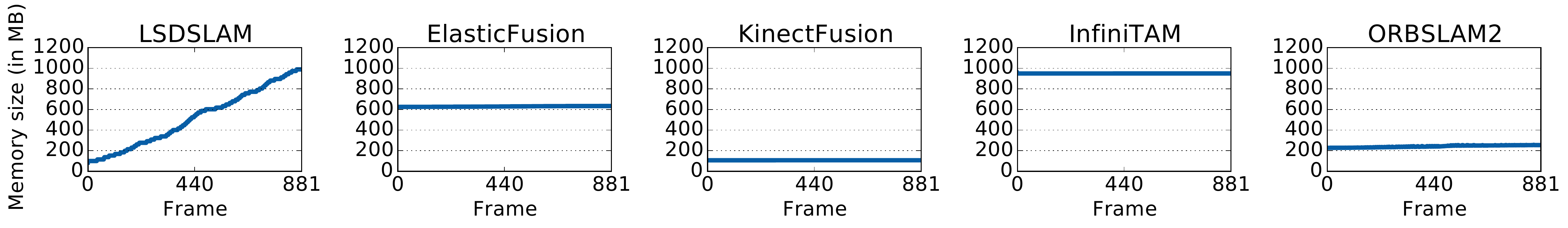}
  \caption{%
    Memory usage of different SLAM algorithms on the trajectory 2 of the ICL-NUIM dataset.
  }
  \label{fig:memory}
\end{figure*}

We evaluated five SLAM algorithms: ElasticFusion, InfiniTAM, KinectFusion, LSD-SLAM, and ORB-SLAM2, over  trajectory 2 of the ICL-NUIM dataset. This trajectory was selected as it successfully tracked on each dataset. The results are shown in Fig.~\ref{fig:memory}.

These results  show  that there are two classes of algorithms depending on whether or not memory is allocated during processing.
For example, KinectFusion allocates a fixed-size volume used for mapping as well as for tracking (through raycast).
This explains why   memory usage is fixed over time.
However, if the camera goes beyond this fixed sized volume, the algorithm will lose track.
In contrast, an algorithm such as LSD-SLAM will continuously build up its map of key frames and so avoid this limitation, but its memory usage will grow as more key frames are discovered.
A more recent algorithm such as ORB-SLAM2 prunes its list of key frames in order to reduce memory consumption. As the results show, this pruning severely reduce memory growth over time.

\subsubsection{Speed/Accuracy Trade-off}
\label{sec:autotuning}

One well known use case of SLAM is in UAV navigation.
Consumer-grade UAVs are increasingly operated indoors or at low altitudes,
and so must navigate through complex spaces and avoid obstacles. This means that highly accurate tracking is required.
The speed of the device also means that the  system must be responsive, and hence  have a high frame-rate. Having a high accuracy and high frame-rate is challenging
and in this experiment  we explore the trade-off between accuracy and frame-rate for different versions of the same algorithm.  

\begin{figure}
  \includegraphics[width=.9\linewidth]{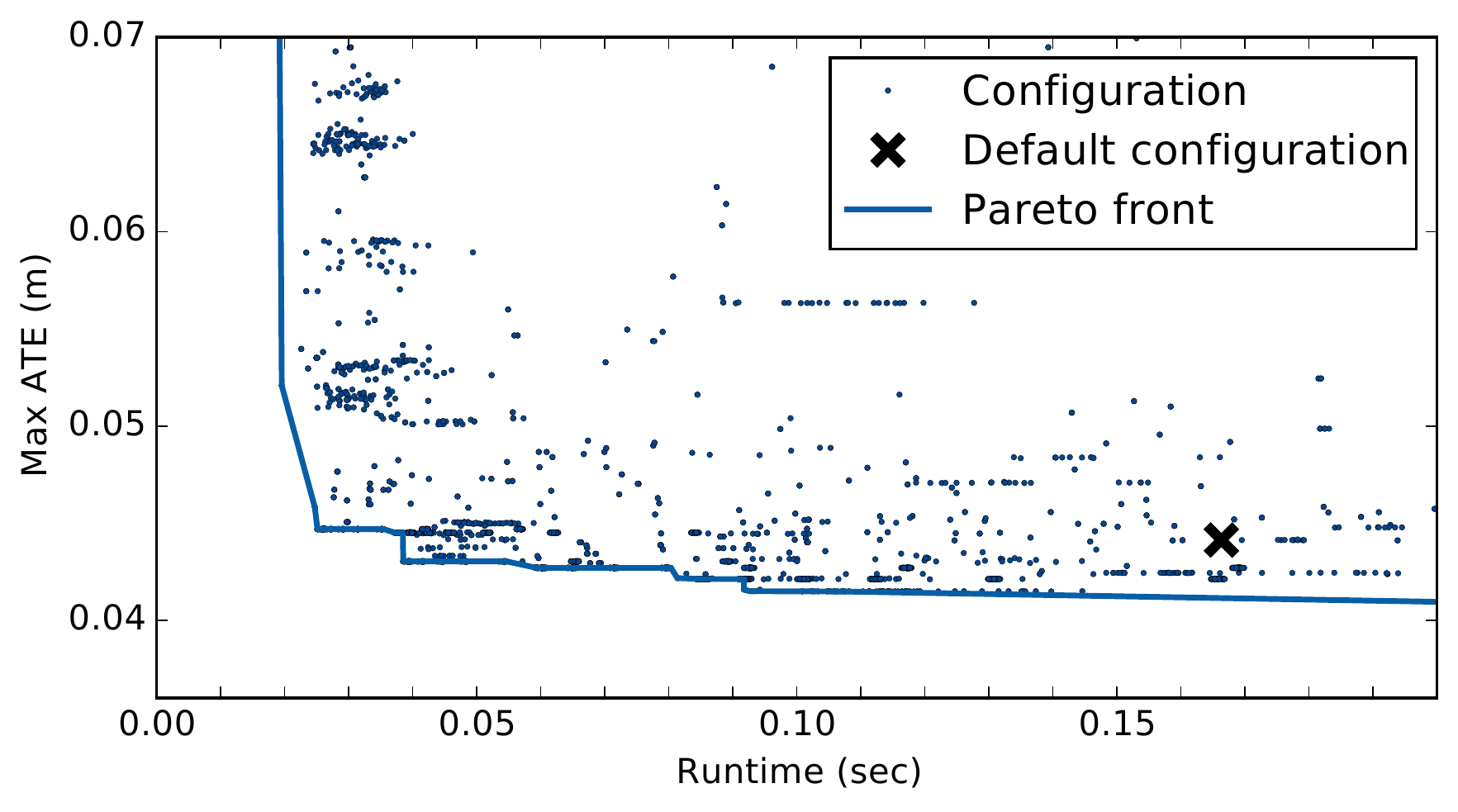}
  \caption{Accuracy/Speed Trade-off exploration performed for KinectFusion (OpenCL) on the ODROID board.}
  \label{fig:dse}
\end{figure}

As our API exposes algorithmic parameters, it is possible to explore the parameter space using smart search algorithms (e.g. HyperMapper~\cite{Bodin2016PACT,nardi2017algorithmic}, OpenTuner~\cite{Ansel2014}).
Figure~\ref{fig:dse} shows the result of exploring the parameter space for the KinectFusion algorithm to find 
accuracy/speed Pareto optimal on the ODROID embedded platform (which could be mounted on a small drone). 

The parameters were ``Volume resolution'', ``$\mu$ distance'', ``Pyramid level iterations'', ``Compute size ratio'', ``Tracking rate'', ``ICP threshold'', and ``Integration rate'' \cite{Nardi2015ICRA}.
The original performance is denoted by the X. By exploring the parameter space, we can see that there are many other parameter settings that lead to either faster or more accurate results.  
In this figure it is clear that by only using different parameters, there is a potential speed-up of 8x compared to the original configuration.
SLAMBench2 allows us to generalise such an experiment over a wider range of datasets and algorithms.

\subsubsection{Productivity Cost of Introducing a New Algorithm}

We evaluate the integration of an algorithm in the SLAMBench2 framework.
Integrating into the UI allows both the use of all supported datasets, and the fair and direct comparison of the algorithm against other integrated algorithms.
Integrating a new algorithm requires implementing a library compatible with the SLAMBench2 API.
For example, we need to implement the \texttt{sb\_new\_slam\_configuration} function, which  declares the algorithm parameters:

\begin{minipage}{\linewidth} %avoid breaking the listing over two pages or columns
\begin{lstlisting}
  bool sb_new_slam_configuration(SBConfig * sbc) {
    sbc->addParameter(TypedParameter<int>(
    "mf", "max-features",
    "Maximum number of features",
    &max_features, &default_mf));
    ....
\end{lstlisting}
\end{minipage}

For ORB-SLAM2, the required functions can be implemented using around 350 lines of code.
This is certainly a straightforward task for the authors of the algorithm, and a simple task for a person familiar with SLAM algorithms. 
Once compiled, this library (i.e. liborbslam2.so) can be directly used with the SLAMBench2 loader:

\begin{lstlisting}
  > sb_loader -i tum_F2_xyz.slam -load liborbslam2.so
  frame ...  orbslam2_ATE
  ...   ...  ...
  3665  ...  0.0088654254
\end{lstlisting}

\subsubsection{Multi-objective Comparison of Multiple SLAM Systems}

\begin{figure}
  \begin{center}
    \textbf{\small~GTX1080}\\
    \includegraphics[width=\linewidth]{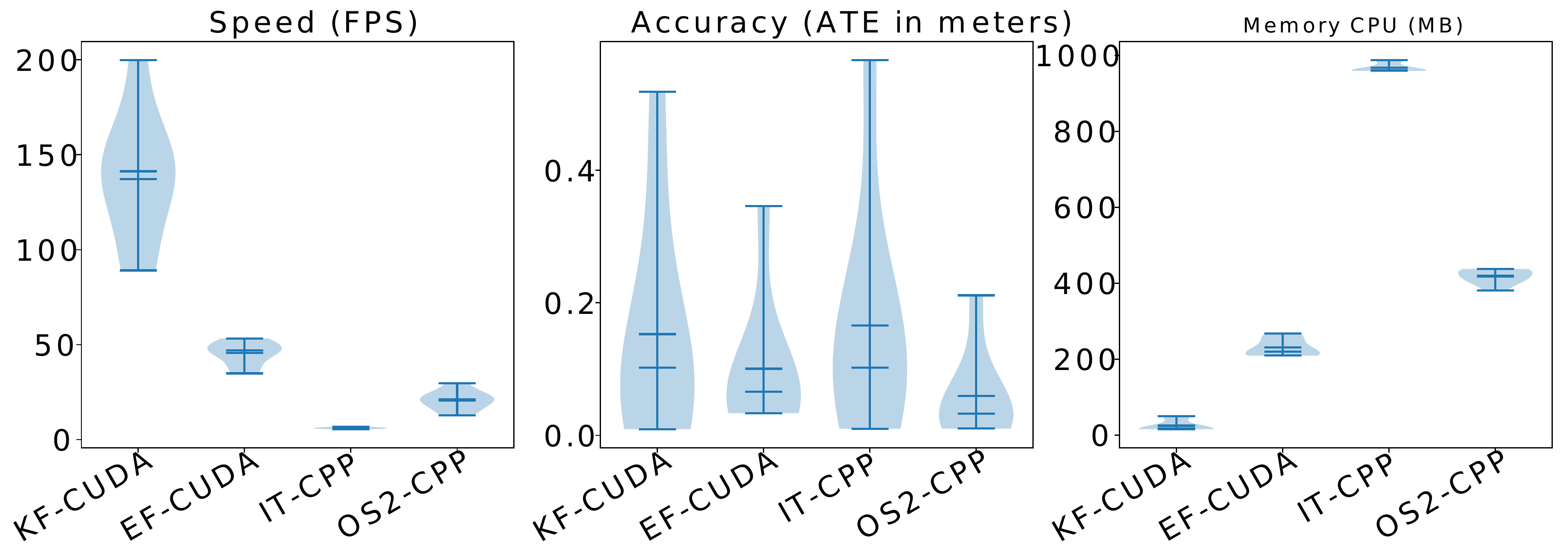}\\
    \textbf{\small~TK1}\\
    \includegraphics[width=\linewidth]{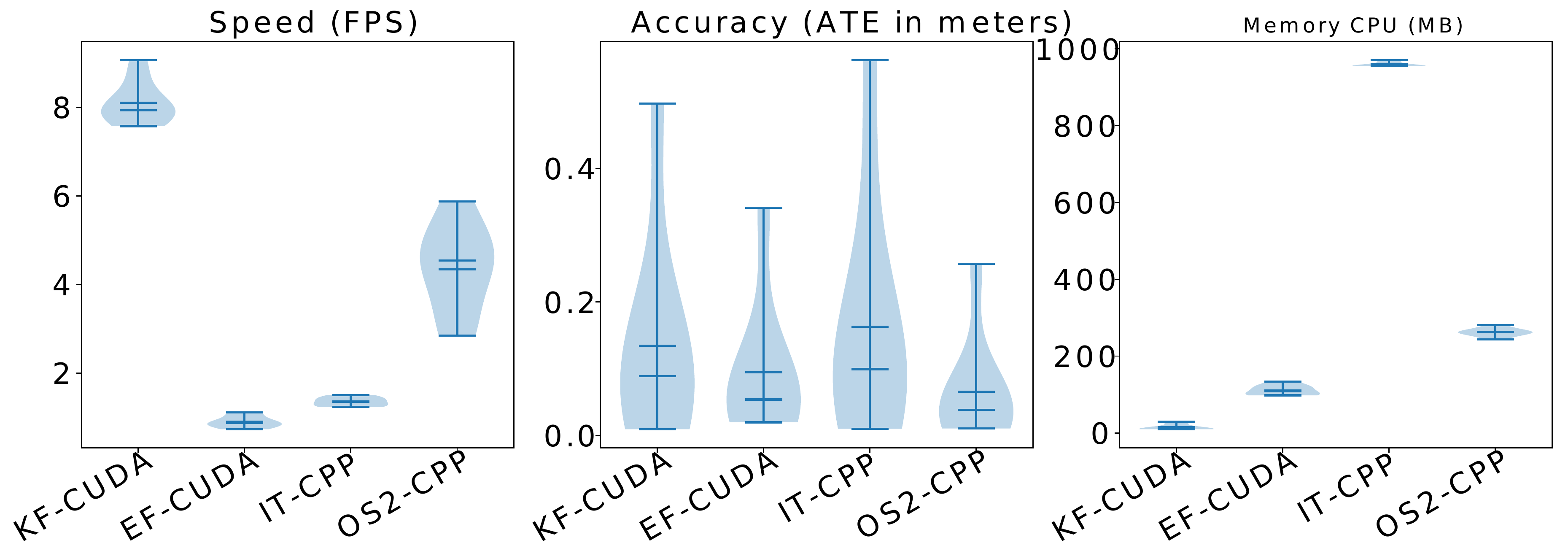}
  \end{center}
  \caption{Comparison of four algorithms run on TK1 and GTX1080, with eight different datasets and over three different metrics. KF: KinectFusion (CUDA), EF: ElasticFusion (CUDA), IT: InfiniTAM (C++), OS2: ORB-SLAM2 (C++). The datasets used were ICL-NUIM (LV0, LV1, LV2, LV3) and TUM (fr1\_xyz,fr1\_rpy,fr2\_xyz,fr2\_rpy).}
  \label{fig:orbslam2}
\end{figure}

Fig.~\ref{fig:orbslam2} shows the result of evaluating ORB-SLAM2, KinectFusion, ElasticFusion, and InfiniTAM on the TK1 and GTX1080 platforms for eight datasets and for three metrics.

The top three diagrams are for the GTX1080 platform, the bottom three are for the
TK1. Each algorithm is labelled along the x-axis and each metric is measured
along the y-axis. The violin of results shows the behaviour of each algorithm over the full data set. 

These violin plots show the variation of metrics over the different datasets. 
For example, the memory consumption of KinectFusion and InfiniTAM are static and thus do not change over the data, while the memory consumption of ORB-SLAM2 is variable even though it is quite reasonably kept in the limited range of 300 MB.
Similarly, the accuracy (Mean ATE) of KinectFusion and InfiniTAM can vary over the datasets from 1 cm up to 60 cm while ElasticFusion seems to be more accurate in general with no more than 30 cm drift.
On the other hand, ElasticFusion is around 50\% slower than KinectFusion.

In this context, ORB-SLAM2 seems to be a good trade-off, keeping its mean ATE down to 30 cm, while being only twice as slow as  KinectFusion on the TK1.
This is significant as ORB-SLAM2 does not use the GPU. 

\section{Conclusions} 

SLAMBench2 provides a significant advantage in terms of evaluating existing SLAM systems and comparing new systems with the state-of-the-art.
The SLAMBench2 methodology has been demonstrated across a wide range of algorithms, datasets, and devices.
Useful insights can be gained. 

We hope that the authors of future SLAM algorithm implementations will find SLAMBench2 to be useful for evaluation, and that they will make their released implementations SLAMBench2-compatible.
So far eight of the most relevant SLAM algorithms are integrated into SLAMBench2, and we hope that this number will increase in the future.
There also are still a large number of datasets that have not been integrated (such as New College~\cite{smith2009new}, SFU Montain~\cite{sfumountain} and Stereo KITTI~\cite{Menze2015CVPR}).
We believe that these datasets will soon be supported, in order to provide reproducible and consistent experimental results when using these artefacts.

We hope to eventually provide something similar to the KITTI benchmarking platform, where SLAM implementations could be uploaded to a cloud service and the SLAMBench2 metrics provided automatically, in order to avoid the effort that is usually required to obtain such results.

In the future, SLAMBench2 could also be extended to allow the integration of more realistic and application-specific evaluation metrics. 
For example, certain classes of algorithm, such as 3D object recognition and scene segmentation, could provide an interesting environment to evaluate the quality of 3D maps generated by SLAM algorithms.

\section*{Acknowledgement}

This research is supported by Engineering and Physical Sciences Research Council (EPSRC), grant reference EP/K008730/1. 

\bibliographystyle{IEEEtran}
\bibliography{main}

\end{document}